# MULTIMODAL SYSTEM FOR SKIN CANCER DETECTION

**V. SYDORSKYI, I. KRASHENYI, O. YAKUBENKO**

**Abstract.** Melanoma detection is vital for early diagnosis and effective treatment. While deep learning models on dermoscopic images have shown promise, they require specialized equipment, limiting their use in broader clinical settings. This study introduces a multi-modal melanoma detection system using conventional photo images, making it more accessible and versatile. Our system integrates image data with tabular metadata, such as patient demographics and lesion characteristics, to improve detection accuracy. It employs a multi-modal neural network combining image and metadata processing and supports a two-step model for cases with or without metadata. A three-stage pipeline further refines predictions by boosting algorithms and enhancing performance. To address the challenges of a highly imbalanced dataset, specific techniques were implemented to ensure robust training. An ablation study evaluated recent vision architectures, boosting algorithms, and loss functions, achieving a peak Partial ROC AUC of 0.18068 (0.2 maximum) and top-15 retrieval sensitivity of 0.78371. Results demonstrate that integrating photo images with metadata in a structured, multi-stage pipeline yields significant performance improvements. This system advances melanoma detection by providing a scalable, equipment-independent solution suitable for diverse healthcare environments, bridging the gap between specialized and general clinical practices.

## INTRODUCTION

Skin cancer is one of the most commonly diagnosed types of cancer, posing a significant public health concern due to its high incidence rates and the risk of severe complications if not detected early [1]. The most effective approach to managing skin cancer is through early detection and prevention [2]. Despite substantial progress in medical imaging and diagnostic technologies, reliably and efficiently detecting melanoma remains challenging. Traditional diagnostic methods depend heavily on dermatologists' expertise, which can be subjective and vary between practitioners. As a result, there is increasing interest in utilizing deep learning techniques to automate and improve the accuracy of skin cancer detection [3]. Progress in dataset curation and related classification challenges has demonstrated potential for fast and accurate skin cancer detection [4]. Deep learning methods have recently gained popularity and have been shown to improve skin cancer detection performance [5–10]. The effectiveness of these models relies heavily on the quality of their training datasets and the limitations of deep learning methods. Additionally, early detection may not always be feasible due to lengthy manual diagnostic procedures [11], and many low-income individuals cannot afford these options. This highlights the need to develop approaches that surpass human diagnostics, providing faster and more accurate results.

Deep learning has become a gold standard for skin cancer classification. Classical approaches using CNN architectures such as ResNets [12], DenseNets [13], and convolutional networks enhanced with attention mechanisms [5] have been widely adopted. More complex pipelines combine segmentation, feature extraction, and attention-based classification neural networks [3]. Synthetic data generated by GANs has proven effective in enhancing model performance [14]. Multi-modal approaches incorporating diverse data modalities further improve classification [15, 16].

Several studies employ hybrid approaches, blending deep learning with traditional machine learning algorithms. For instance, [17] uses VGG16 for feature extraction, followed by XGBoost for final image classification. This study also leverages synthetic data as part of its data augmentation strategy. Another noteworthy approach involves skin cancer detection using genetic data. In [18], various machine learning algorithms, including KNN, SVM, and XGBoost, are applied to classify melanoma. Similarly, [19] explores the use of NIR spectroscopy as input data, utilizing XGBoost, LightGBM, 1D-CNNs, and other machine and deep learning methods for classification.

Despite progress, several research gaps remain:

- Development of complex multi-modal systems integrating different modalities (e.g., image and tabular data) in parallel or sequential architectures.
- Optimization of multi-modal neural networks for datasets with and without meta-features.
- Adaptation to challenging imaging conditions, such as images captured using mobile devices, addressing data imbalance and quality variations.

In this study, we propose a novel framework for melanoma detection that integrates visual, structural, and lesion metadata with patient information such as age and sex. Our solution combines multi-modal neural networks for processing visual and metadata inputs, and a gradient-boosting model for metadata analysis unified within a three-stage pipeline. Additionally, we introduce a two-step training methodology to accommodate datasets with varying metadata availability. Advanced training techniques and feature engineering are applied to address class imbalance, ensuring robust and efficient melanoma detection. Finally, we enhance system performance through two stages of feature engineering, enabling robust and efficient melanoma detection.

## MATERIALS AND METHODS

**Data. This research utilizes several data sources:**

- Data from ISIC 2024 Kaggle Challenge [20] - Main Data.
- Data from ISIC Archive [21] - ISIC Archive Data.
- Artificially generated image data [22] - Generated Data.

The primary dataset from the ISIC 2024 Kaggle Challenge is the foundation for most of the training and validation processes. However, additional data sources play a crucial role in enhancing the proposed system, particularly in improving the neural network performance. However, diverse datasets introduce domain shifts and varying feature subsets, creating challenges in harmonizing and effectively integrating the information. The proper fusion of these heterogeneous data sources represents a key contribution to our work, enabling robust performance across different data modalities and domains.

**Competition Data.** The ISIC 2024 Kaggle Challenge dataset [20] includes images and metadata from single-lesion crops extracted from 3D total body photos (TBP) [23]. This dataset presents several challenges:

- Lower data quality compared to dermatoscopy images. The images resemble close-up smartphone photos, making them highly relevant for telehealth applications, where patients often submit similar-quality images (Fig. 1).
- Different labeling confidence. The dataset includes two categories of labels: "strongly-labeled tiles," verified through histopathology, and "weakly-labeled tiles," which were not biopsied and considered benign by a doctor's assessment.
- Severe class imbalance. The dataset contained 401059 tiles, where 400666 (99.902 %) are benign, and 393 (0.098 %) are malignant.

Each image represents 15x15 mm of skin area but may come with a slightly varying resolution centered around 133 pixels (Fig. 2). Besides image data, the dataset includes metadata about patients, tile location, image characteristics, and extracted features derived from and extracted features derived from [24] and [25].

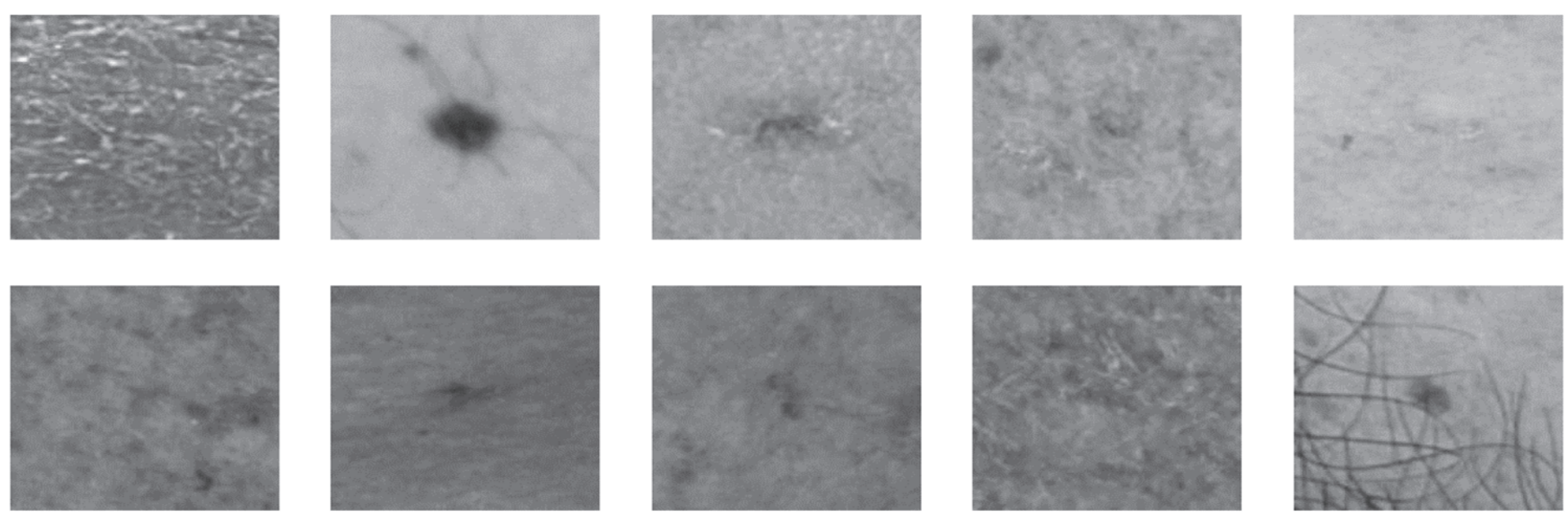

*Fig. 1.* 1st row – benign; 2nd row – malignant images from ISIC 2024 Kaggle Challenge

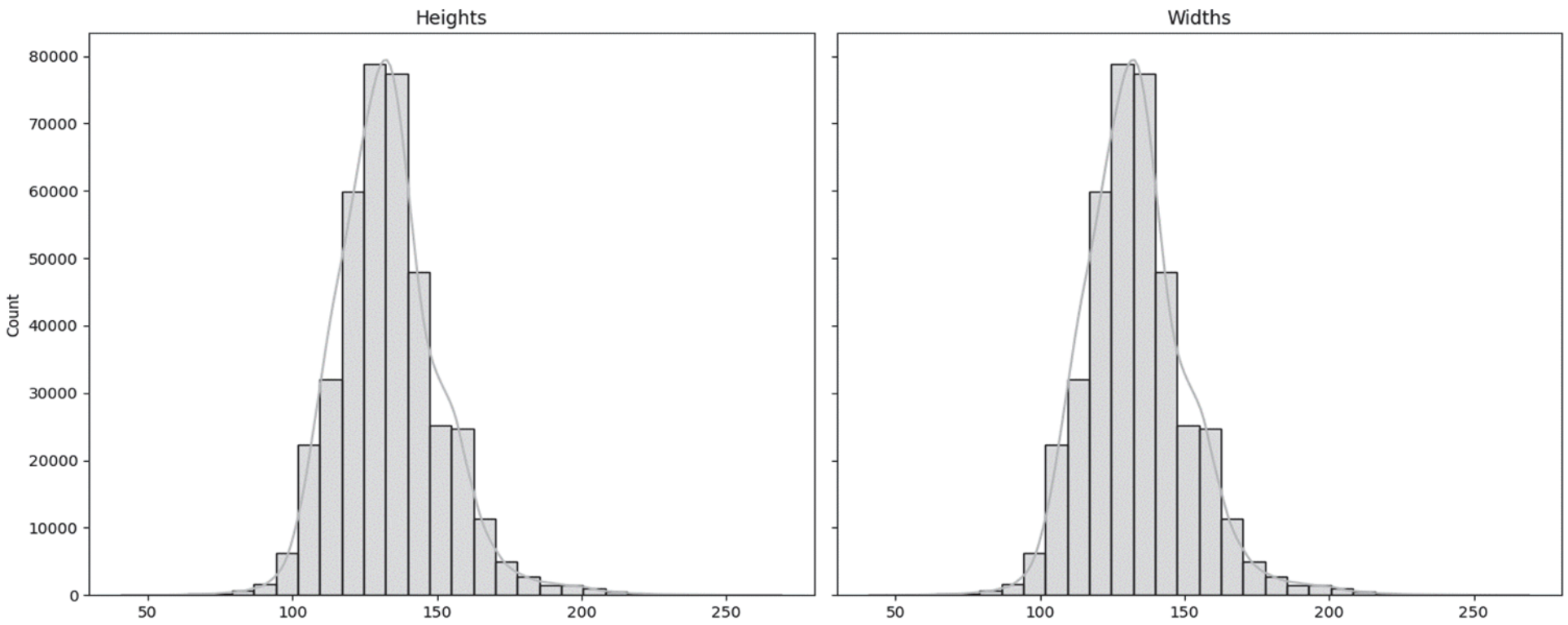

*Fig. 2.* Distribution of image shapes in Data from ISIC 2024 Kaggle Challenge

**ISIC Archive Data.** The ISIC Archive dataset [21] contains 81.722 images accompanied by metadata. However, the dataset is highly unstructured due to its compilation from various data sources and competitions. For this study, most of the available meta-features are disregarded, and only patient information, target labels, and images are utilized for system development.

To avoid potential data leakage, all patients included in the ISIC 2024 Kaggle Challenge dataset [20] are excluded from the ISIC Archive data. Additionally, images lacking explicit benign/malignant labels are removed. After such a filtration ISIC Archive contains 71080 images, where 61910 (87.099 %) are benign, and 9170 (12.901 %) are malignant. This dataset has approximately 5.6 times fewer total images than [20], but it contains 23.3 times more malignant images.

The primary challenge with the ISIC Archive dataset lies in its substantial data diversity. It aggregates skin lesion tiles from various sources, including dermatoscopy and standard photo images. The images differ significantly in size, aspect ratio, scale, and padding, influenced by the medical devices used to capture them (Fig. 3). Most of the images are dermatoscopy and come into higher resolution – median height resolution is 3024, and width is 2016 (Fig. 4). This diversity introduces a significant domain shift compared to the ISIC 2024 Kaggle Challenge dataset [20]. Despite these challenges, the ISIC Archive dataset is a valuable source of malignant images, addressing their severe undersampling in the primary dataset. Its inclusion enriches the data diversity, improving model robustness and generalization.

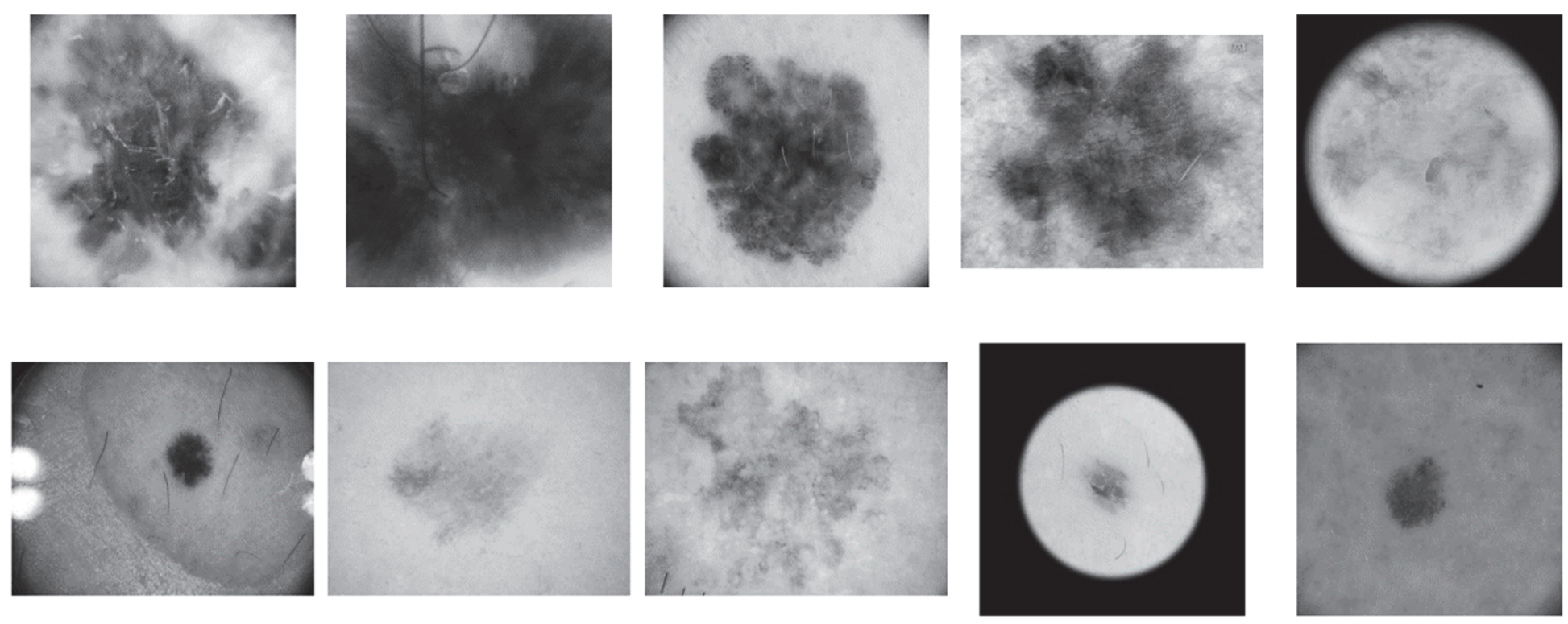

*Fig. 3.* 1st row – benign; 2nd row – malignant images from ISIC Archive

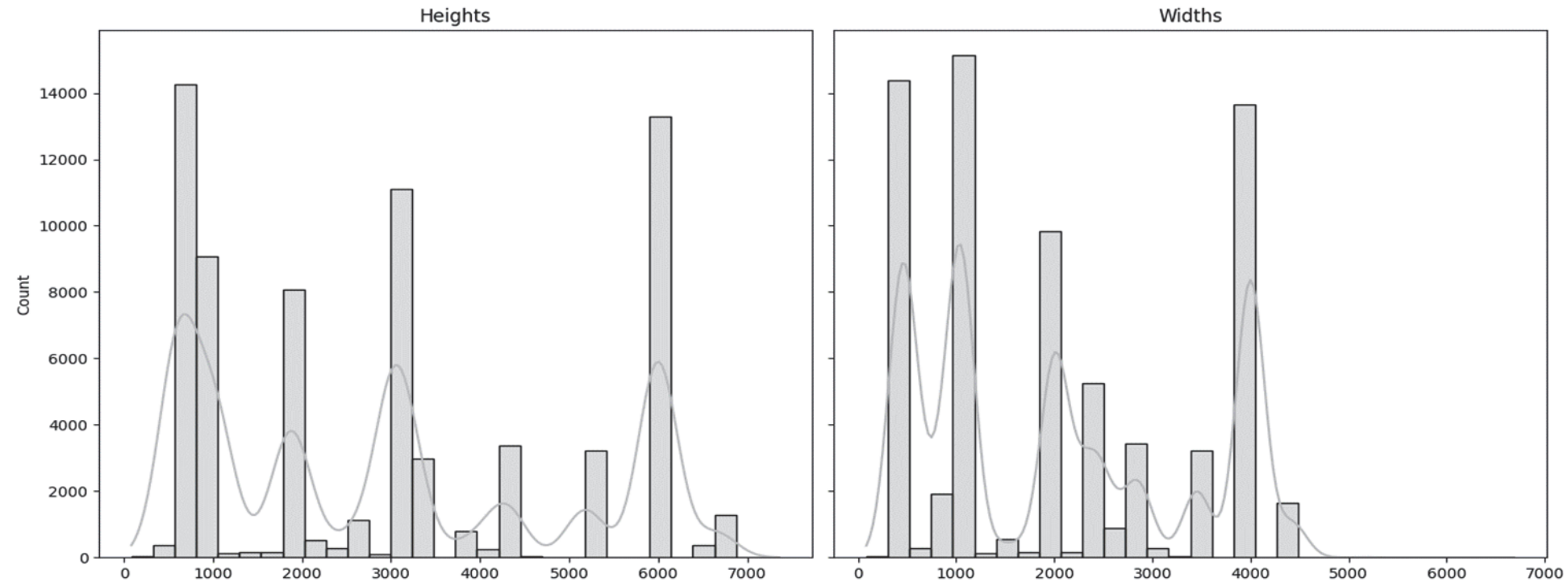

*Fig. 4.* Distribution of image shapes in ISIC Archive

**Generated Data.** The generated image dataset from [22] was created using the Stable Diffusion 2 model [26]. It consists of 6.012 images, with a nearly equal distribution of classes: 3.012 malignant and 3.000 benign images. All images are standardized to a resolution of 512×512 pixels. While the generated images can often be distinguished by artifacts and smooth textures characteristic of generative models (Fig. 5), the model performs well in preserving malignant lesions' shapes and visual characteristics. This fidelity provides valuable information that can enhance the training of deep learning models by supplementing the limited data of malignant cases in real-world datasets.

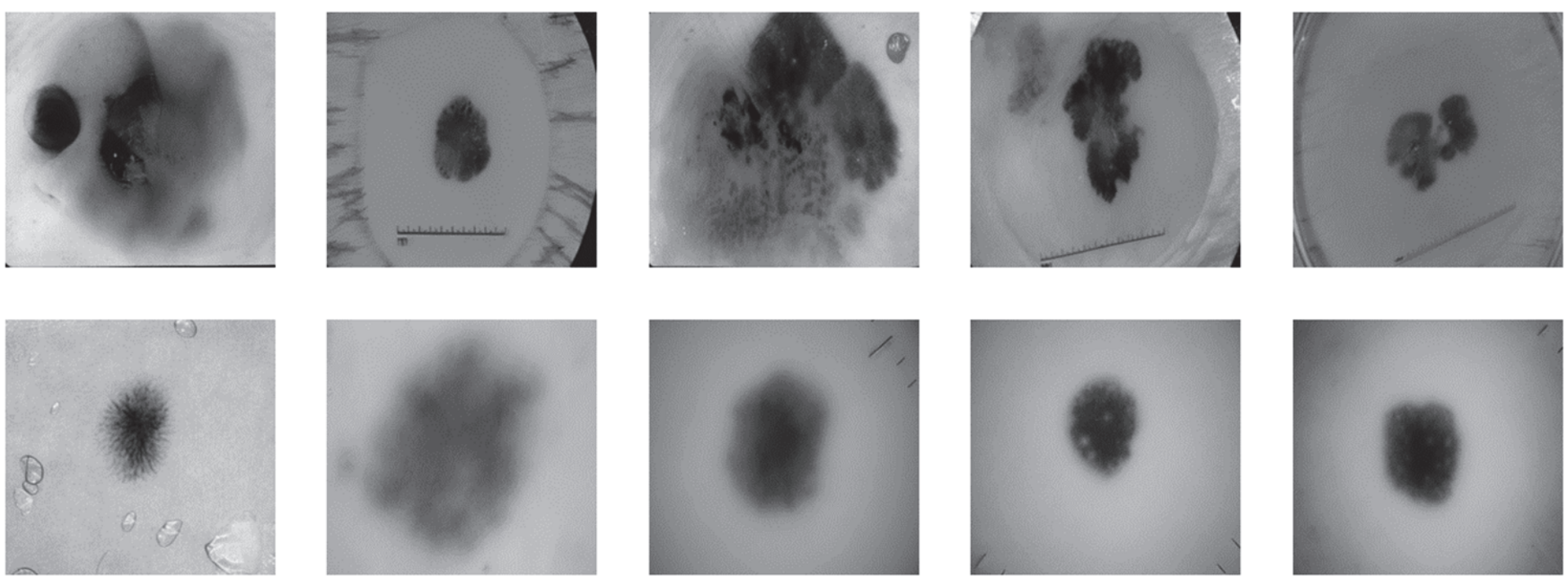

*Fig. 5.* 1st row – benign; 2nd row – malignant images from Generated Dataset

## METHODS

This subsection describes the proposed system, including its components, models, optimization processes, evaluation metrics, and validation procedures.

**Metrics.** To evaluate the proposed system and models, the following metrics are utilized:

- ROC AUC. A standard metric to measure the overall performance of a binary classification model.
- Partial ROC AUC [27]. This metric calculates the area under the ROC curve only for True Positive Rates (TPR) above 80%. The score ranges from 0.0 to 0.2, emphasizing performance in high-sensitivity regions critical for clinical applications.
- Top-15 retrieval sensitivity [28]. This metric is the most appropriate to real clinic scenarios when a dermatologist has limited time for a patient and should pay attention to the most suspicious lesions [29].

Out of fold (OOF), Mean Fold metrics will be reported.

**Validation.** To evaluate the models, a classical 5-fold cross-validation approach is used. Folds are stratified based on the target label (benign/malignant), and no patient overlaps across folds are ensured.

In cases where a two-stage training approach is employed (first stage: ISIC Archive + Main + Generated Data; second stage: only Main Data), the datasets are split separately, and respective folds are merged afterward.

For models using only tabular data, validation is repeated five times with different random seeds, and average scores are reported. Hyperparameter tuning for tabular models is performed using the Optuna algorithm [30], with the tuning strategy discussed in the Tabular Approach.

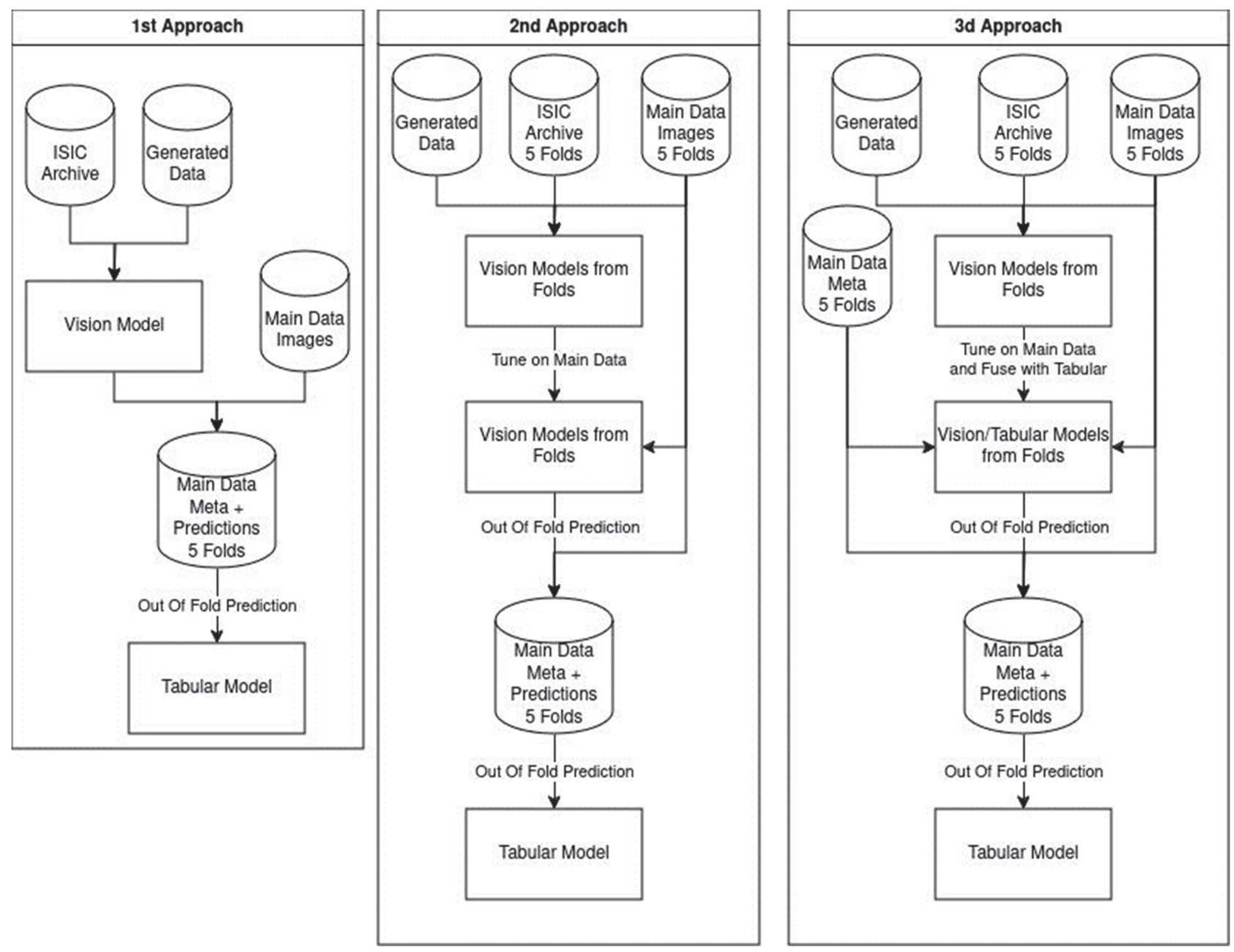

*Fig. 6.* Models pipelines and validation schemes

The two-stage system, which integrates Vision and Tabular models, can follow several aggregation approaches (Fig. 6):

1. Vision-Only Pretraining:
   - Train the Vision model on ISIC Archive and Generated data.
   - Generate predictions for the Main dataset.
   - Train the Tabular model using Vision model predictions and tabular features.
2. Vision Model Pretraining and Fine-Tuning:
   - Train the Vision model on ISIC Archive and Generated + Main datasets.
   - Fine-tune the Vision model on the Main dataset only.
   - Generate out-of-fold predictions on the Main dataset.
   - Train the Tabular model using these predictions and tabular features.
3. Multi-Modal Pretraining and Fine-Tuning: Same as Approach 2, but tabular data is also incorporated during the Vision model fine-tuning.

In the final third stage of the system (Fig. 7), the Optuna algorithm is used for the final stage of coefficient optimization. However, it is crucial to recognize that the validation metrics obtained during the second stage, particularly for the second and third approaches, may be unreliable and could lead to overly optimistic outcomes. Similarly, the final stage lacks validation, which can further contribute to unfavorable results. To address these limitations and ensure a robust evaluation of the final system's performance, the Public and Private Leaderboards from the Kaggle Competition [31] are used as benchmarks. The Public test set contains approximately 140.000 tiles, while the Private test set includes around 360.000 tiles. These external benchmarks provide a more realistic and unbiased assessment of the system's capabilities.

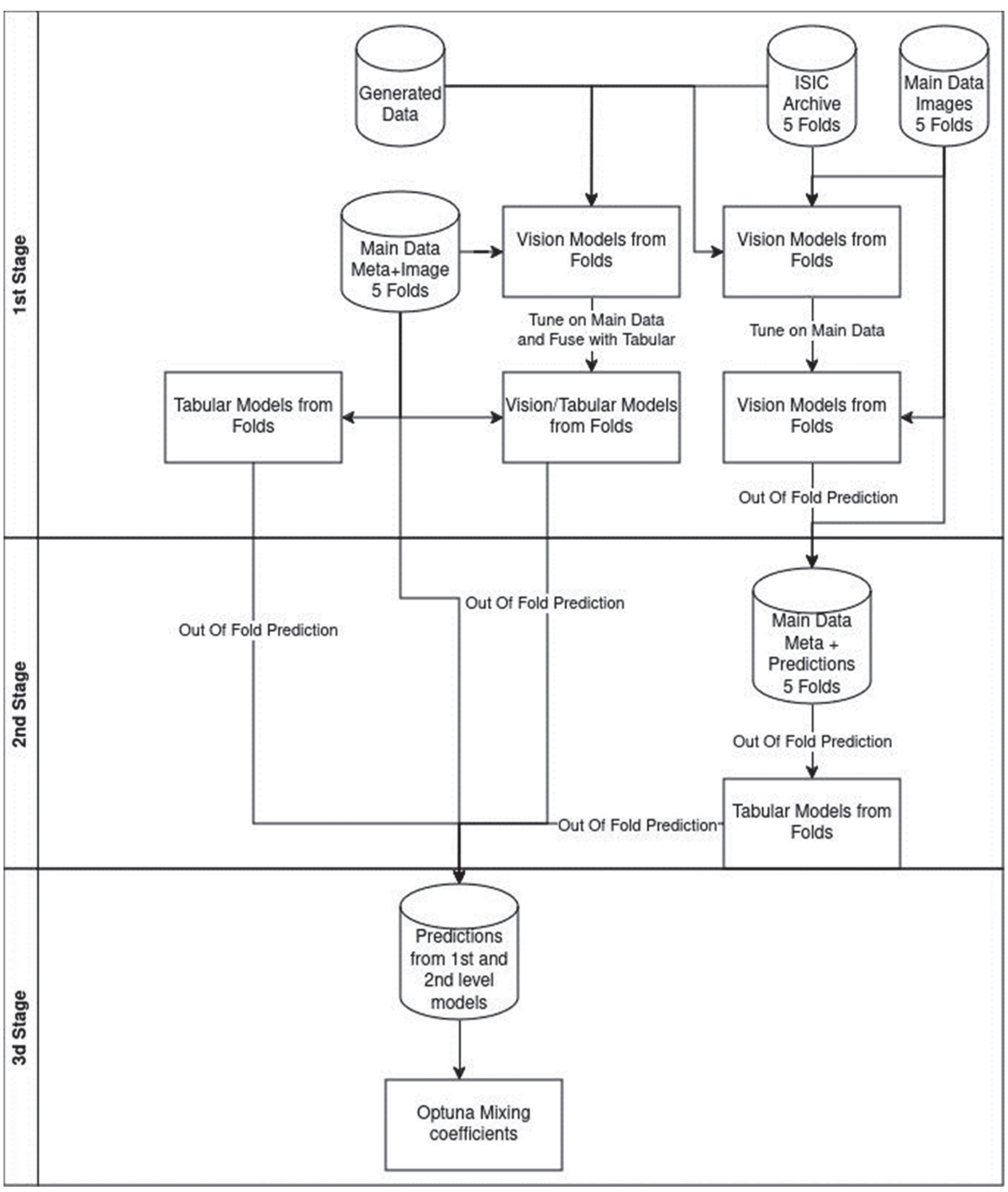

*Fig. 7.* Scheme of Multi-model fusion system

**Feature Engineering.** Basic preprocessing steps are applied to the tabular data, including handling missing values and removing redundant features. Specifically, missing numerical values are filled with the median, accompanied by adding a missing indicator feature, while missing categorical values are replaced with a new “nan” category. Columns that are static or present only in the training data are dropped to ensure consistency across datasets.

In addition to basic tabular features, several advanced features are manually engineered to capture spatial, color, and physical relationships inherent in the data. Initial proposals for these features are generated using ChatGPT [32] and refined through pruning. Key engineered features include:

- Lesion size ratio: The ratio of the minimum to the maximum diameter of the lesion.
- Hue contrast: The difference in hue between the lesion’s center and periphery.
- Perimeter-to-area ratio: The ratio of the lesion’s perimeter to its area.

A critical aspect involves the comparison of lesion characteristics within the same patient or body region, motivated by [33]. To address this, aggregation features are introduced:

1. Deviation within patients: StandardScaler [34] is applied within each patient-id group to capture deviations relative to other lesions of the same patient.
2. Deviation within body regions: StandardScaler is applied within combined patient-id and anatomic-site-general groups, reflecting deviations in specific body regions (e.g., arm, leg).
3. Extremes within patients: Maximum and minimum feature values per patient are calculated. Given the limited patient sample size, these features are discretized using QuantileTransformer [35] to mitigate overfitting risks.

Additionally, incorporating skin type as a feature inspired by [36] improved performance.

Finally, categorical features are one-hot encoded to prepare the data for model training.

**Multi-Modal Neural Net: Image + Tabular data.** Multi-Modal Vision + Tabular model is trained in 2 stages:

1. A CNN Encoder combined with a Multilayer Classifier is trained on the ISIC Archive, Generated, and Main datasets.
2. The pre-trained CNN Encoder is combined with a randomly initialized Feed-Forward Tabular Neural Net in a new Multilayer Classifier, and this combined model is fine-tuned only on the Main dataset.

All images are resized to a resolution of 128×128. Images from the ISIC Archive are also center-cropped before resizing. Continuous tabular features are normalized using the StandardScaler, while categorical features are one-hot encoded.

ConvNeXt V2 Pico [37], EdgeNeXt Base [38], and EfficientNetV2 B0 [39] CNN architectures are used. Pre-trained models from the timm repository [40] are starting points for first-stage training: convnextv2_pico.fcmae_ft_in1k, edgenext_base.in21k_ft_in1k, tf_efficientnetv2_b0.in1k. EdgeNeXt Base is used in one of the multi-modal architectures to leverage its attention mechanisms, prioritizing robustness over inference speed. EfficientNetV2 B0 is employed as a first-level model to generate predictions for the second-level pipeline, benefiting from its high inference speed. ConvNeXt V2 Pico balances inference speed and accuracy, making it suitable for prediction generation and multi-modal architectures.

Heavy augmentations are applied to enhance model robustness and mitigate overfitting in undersampled malignant cases. These include various spatial, color, blurring, distortion, and dropout augmentations, introducing variability and improving generalization (see Section Detailed Neural Net Architecture and Training Setup).

To address class imbalance during training:

- A balanced sampling strategy is employed in the first stage, ensuring equal representation of positive and negative classes.
- A “square” balancing strategy is applied in the second stage to refine class distribution further.

For generating predictions, the last and best (based on validation Partial ROC AUC) is used. Predictions are averaged across test-time augmentations (TTA), incorporating four flips to increase accuracy and robustness.

**Detailed Neural Net Architecture and Training Setup.** Images are first normalized to the [0, 1] range and then normalized to ImageNet statistics [41]. For image resizing, methods from the OpenCV library [42] are used, specifically INTER_AREA and INTER_LANCZOS4. The final choice is INTER LANCZOS4, though the overall difference between methods is marginal.

Both first and second-stage models are trained with the following series of augmentations:

- Transpose with a probability of 0.5.
- Vertical Flip with a probability of 0.5.
- Horizontal Flip with a probability of 0.5.
- Random Brightness and Contrast adjustment, with changes in brightness and contrast within the range [–0.2, 0.2] and a probability of 0.75.
- One of the following blurs: Motion, Median, Gaussian, or Gaussian Noise, with variation in the range [5, 30] and a blur kernel size limit of up to 5, applied with a probability of 0.7.
- One of the following distortions: Optical (limit up to 1.0), Grid (5 steps with a limit of up to 1.0), or Elastic Transform (alpha up to 3), applied with a probability of 0.7.
- CLAHE with a clip limit of up to 4, applied with a probability of 0.7.
- Random adjustment of hue, saturation, and value. Hue changes within [-10, 10], saturation within [–20, 20], and value shift within [–10, 10], with a probability of 0.5.
- Shifting, scaling, and rotation of the image. Shift within [–0.1, 0.1], scale within [–0.1, 0.1], and rotation within [–15, 15] degrees, applied with a probability of 0.85.
- Coarse Dropout with one hole of 48 width and height, applied with a probability of 0.7.

Models for both stages are trained with a batch size 64 and Binary cross-entropy loss. First-stage models are trained for 10 epochs, while second-stage models are trained for one epoch. The limited number of epochs is due to the large dataset size combined with a small number of positive samples. Balanced sampling is used, and overfitting tends to occur early.

For optimization, the Adam optimizer [43] is used in the first stage, while RAdam [44] is used in the second stage. The learning rate settings are as follows:

- For the first stage, the starting learning rate is set to. For the EdgeNeXt model, weight decay is set to.
- For the second stage:
- The CNN encoder (from the first stage) starts with a learning rate of 1e-4.
- The Feed-forward Tabular Neural Net and Multilayer Classifier start with a learning rate of 1e-3.

Such a choice of different starting learning rates is crucial to fit the entire model within one epoch while avoiding overfitting the training data or one of the modalities (image or tabular). Learning rates are reduced using cosine scheduling to 10 times smaller than their initial value. For second-stage training, the last checkpoint from the first stage is used. “Square” balancing weights for second-stage training are illustrated in Equation below. For EfficientNetV2, the starting learning rate is set to and reduced to. This model is trained on all ISIC Archive and Generated data, resulting in good convergence.

$$class\ weight_k = \sqrt{\frac{\sum_{j=1}^{C} \sum_{i=1}^{N} class_{i,j}}{\sum_{i=1}^{N} class_{i,k}}}$$

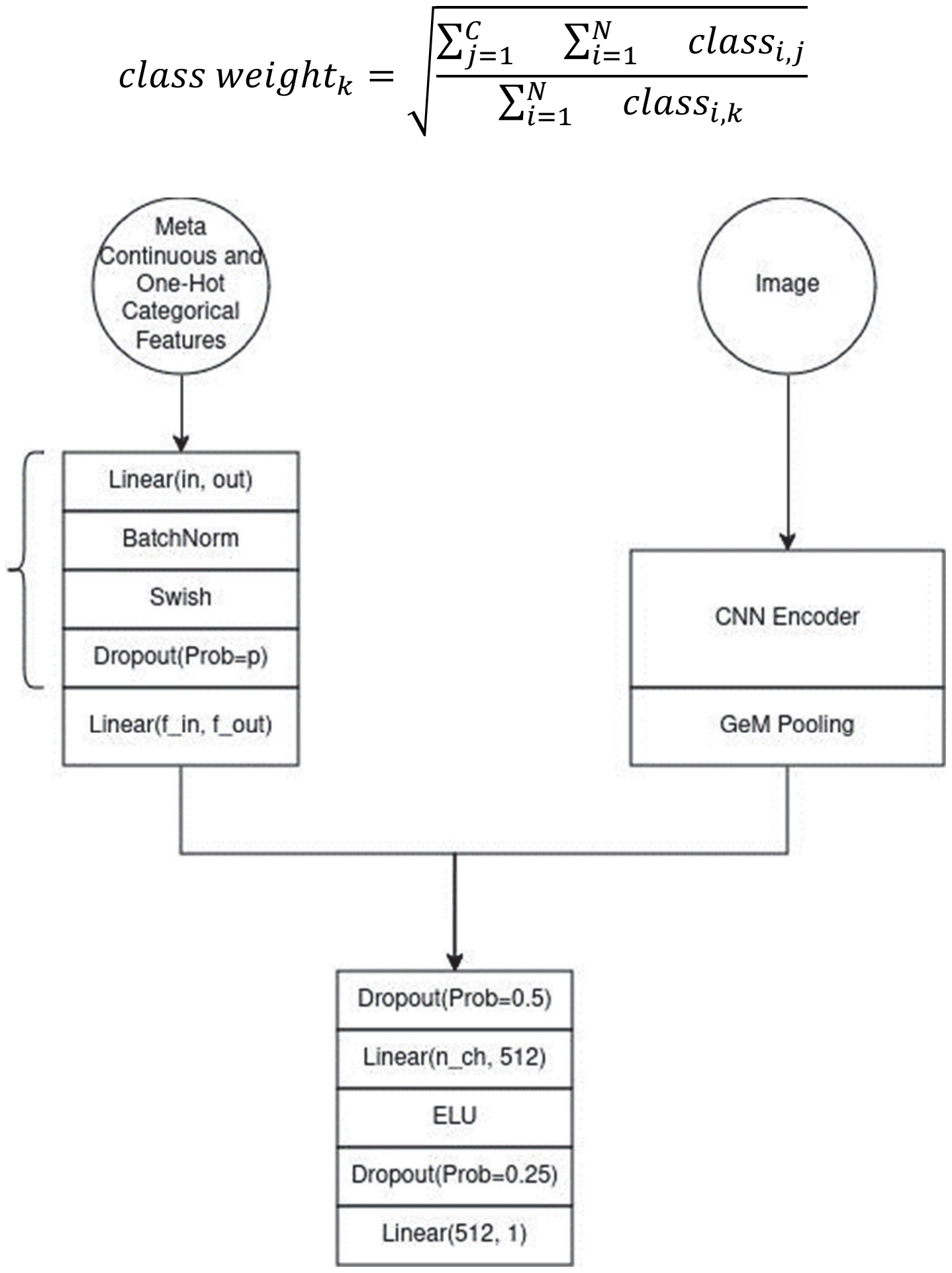

*Fig. 8.* Vision and Tabular general model architecture

The overall Image and Tabular model architecture is inspired by [16]. The resulting model architecture is shown in Fig. 8. Detailed values for the number of hidden channels and other hyperparameters can be found in Tables 1 and 2.

**Table 1.** CNN Encoders

| CNN Encoder | Embedding Shape | Vision and Tabular Embedding Shape |
|---|---|---|
| ConvNeXt | 512 | 576 |
| EdgeNeXt | 584 | 648 |
| EfficientNetV2 | 192 | – |

**Table 2.** Feed Forward Net for Meta Features

| Layer | In Channels | Out Channels | Dropout Probability |
|---|---|---|---|
| 1 | 200 | 256 | 0.3 |
| 2 | 256 | 512 | 0.3 |
| 3 | 512 | 128 | 0.3 |
| Final | 128 | 64 | – |

An important feature of the multi-modal model is a substantial bottleneck in the final Feed-forward Net layer. This bottleneck is critical for preventing overfitting to the tabular branch. Using a more significant number of channels results in faster overfitting and poorer final results.

**Tabular Approach.** To address the class imbalance and enhance the efficiency of training and hyperparameter selection, RandomUnderSampler [45] is used as the initial step for all tabular models. The underlying model used is a boosting algorithm (LightGBM [46] or XGBoost [47]), with or without early stopping.

- For boosting without early stopping, a single model is trained on the entire training dataset, and the number of epochs is selected as one of the hyperparameters.
- In the case of boosting with early stopping, an ensemble of five models is trained. Each model is fitted on 4/5 of the training data, with early stopping performed on the remaining 1/5 (the data split follows the same principle described in the validation section) (Fig. 9).

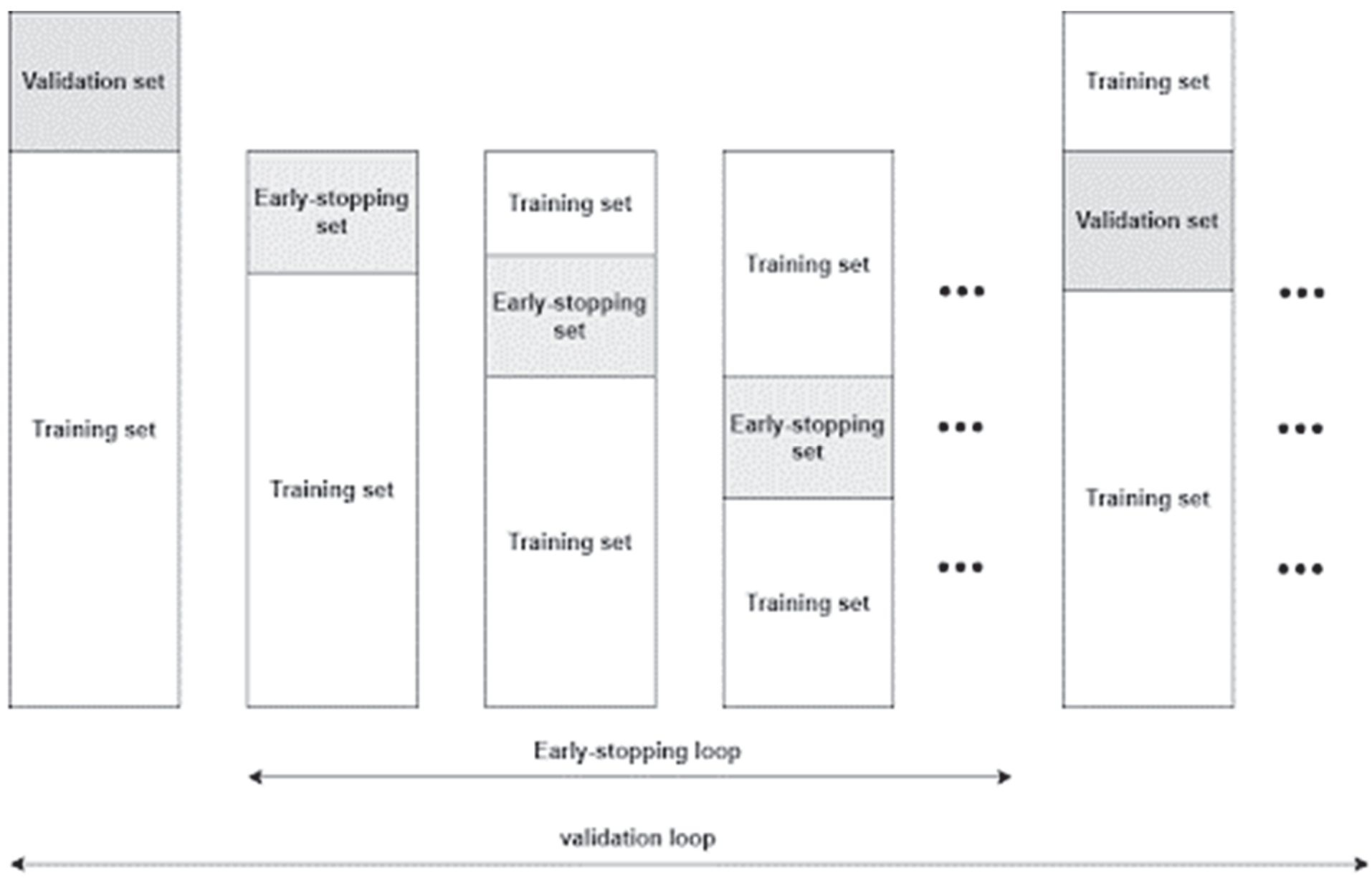

*Fig. 9.* Validation strategy for Tabular Models

Hyperparameter tuning, including the under-sampling ratio and the number of epochs for models without early stopping, is conducted using the Optuna optimization framework in the following steps:

1. Initial Optimization: 300 Optuna trials are performed without predefined starting parameters.

2. Parameter Refinement: The best parameters from the top five trials are combined to reduce overfitting and ensure a more robust solution. For numerical parameters, medians are calculated across the top-performing trials. For categorical parameters, the most frequent value or the value from the model with the highest Partial ROC AUC score is selected.

The primary fusion scheme is depicted in Fig. 7.

1. In the first stage, tabular and multi-modal tabular/image neural models are trained. An image-only neural model is also trained for use in the second stage.

2. In the second stage, a tabular model is trained using tabular features and the outputs of the image-only neural model.

3. In the third stage, the outputs of all models from the first stage (except the image-only neural model) and the outputs of the second stage models are ensembled using Optuna coefficient optimization. The optimization is performed on Partial ROC AUC directly on validation folds.

Predictions are generated from each fold model in the final system, increasing inference time but significantly improving robustness. This tradeoff is crucial for real-world applications such as skin cancer detection.

Due to the different nature of models and variations in the number of positive samples, the final probability distributions for each model can vary (Fig. 10). The distributions are standardized using the rank method to address this, where probabilities are converted to ranks before ensembling.

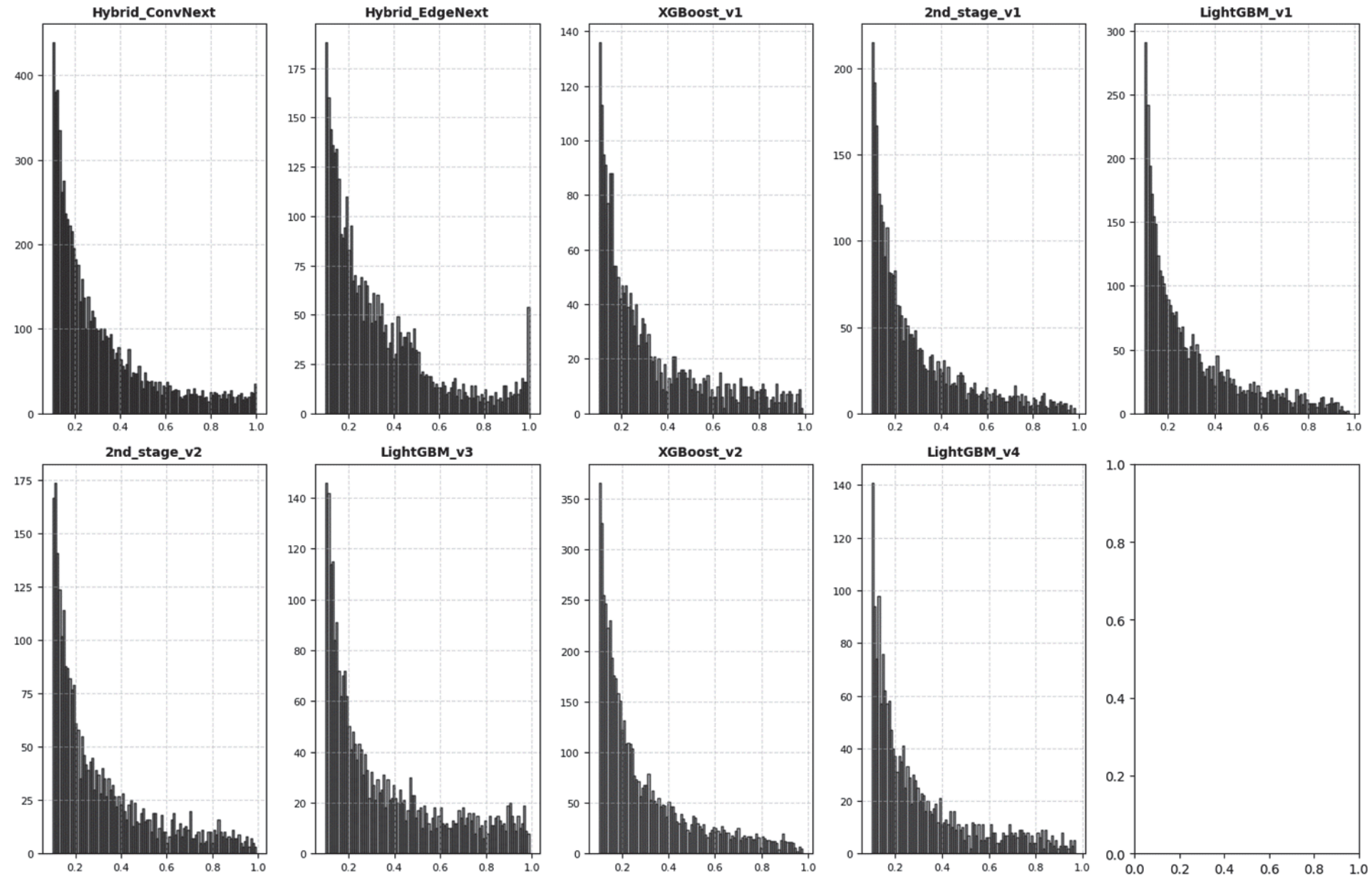

*Fig. 10.* Probability distribution of different models, trimmed by 0.1. Trimming is needed because most of the probabilities are lower than 0.1, while the most interesting part is above this point

For the final Optuna optimization stage, overfitting to the training (validation) set is a notable risk. The top 10 results from 5 optimization runs are collected and averaged to mitigate this. The results of the coefficient search are shown in Fig. 11. Final ensemble weights are adjusted manually to refine the system further.

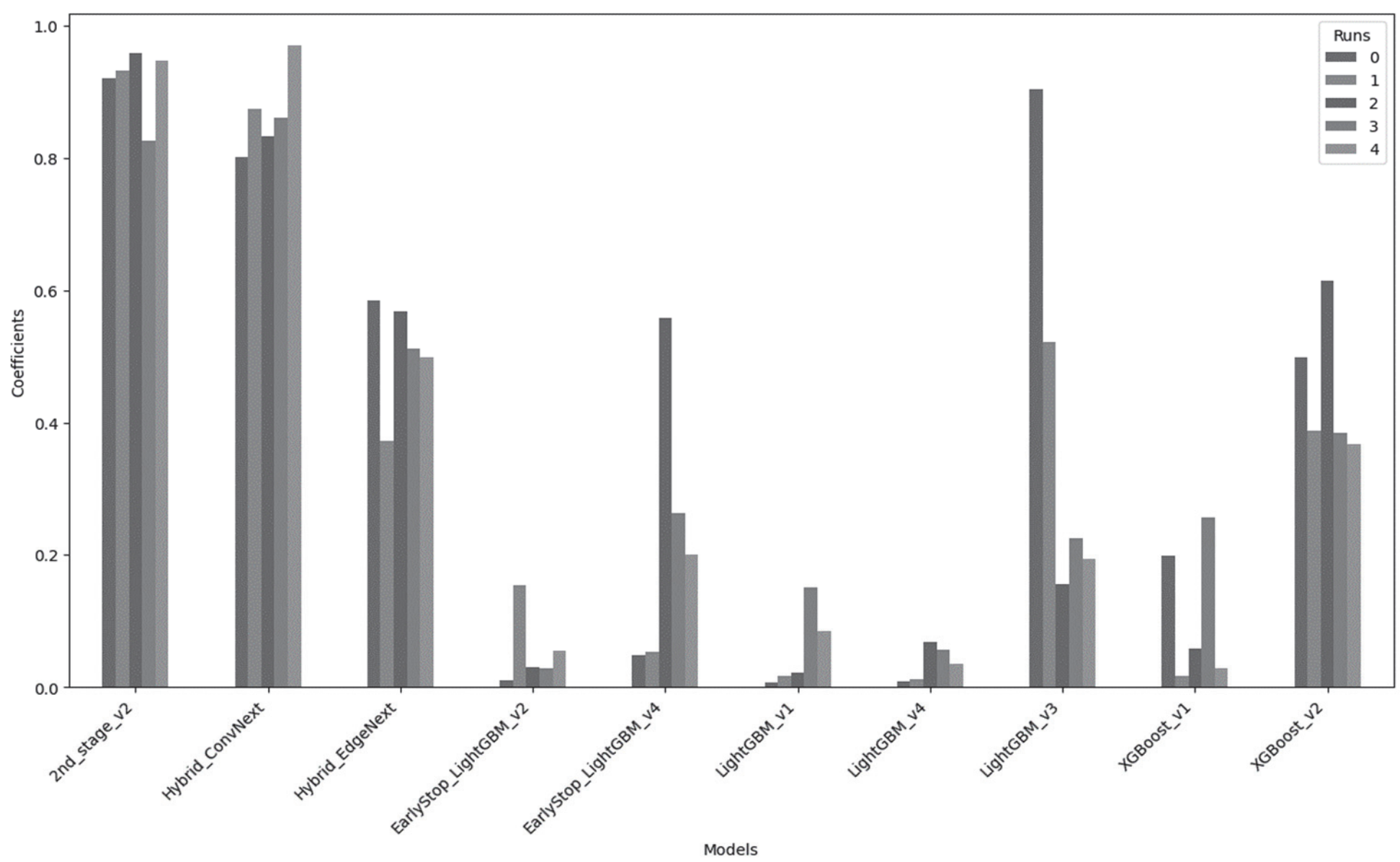

*Fig. 11.* Coefficients of 3rd Stage (Three-stage v2) obtained from Optuna across several runs

## RESULTS

In the results, Tables 3, 4, 5 and the coefficient Table 12 different versions refer to the following:

- Different versions of Two-stage models. These correspond to the different approaches illustrated in Fig. 6.
- Different versions of XGBoost and LightGBM models. These reflect minor adjustments in the Optuna configurations or feature setups. For instance, version $\geq 2$ incorporates the skin tone feature.

We evaluate the proposed methods using validation, private, and public datasets. The primary metrics include Partial ROC AUC, ROC AUC, and Top 15 Retrieval Sensitivity. For the validation dataset, we report both Out-of-Fold (OOF) and Mean metrics, with their relevance discussed in Section Metrics. Due to limitations, we only report the Top 15 Retrieval Sensitivity for the validation dataset.

### Metrics of One-stage and Two-stage Models

The proposed two-stage v2 model demonstrates superior performance across all datasets in Partial and Full ROC AUC metrics, as detailed in Tables 3 and 4. The Multi-Modal ConvNext model achieves the highest Top 15 Retrieval Sensitivity (Table 5). While validation results for two-stage models may appear over-optimistic (see Section Validation), their consistent outperformance on public and private datasets supports such conclusions.

**Table 3.** Partial ROC AUC of One-Stage and Two-Stage Models

| Model | OOF | Mean | Private | Public |
|---|---|---|---|---|
| Two-stage v2 | 0.17666 | 0.17862 | 0.16941 | 0.18608 |
| Multi-Modal ConvNext | 0.17497 | 0.17698 | 0.16090 | 0.17714 |
| XGBoost v2 | 0.17348 | 0.17460 | – | – |
| XGBoost v1 | 0.17252 | 0.17351 | – | – |
| EarlyStop LightGBM v4 | 0.17225 | 0.17305 | – | – |
| LightGBM v3 | 0.17173 | 0.17266 | 0.16107 | 0.18400 |
| LightGBM v4 | 0.17105 | 0.17210 | – | – |
| EarlyStop LightGBM v1 | 0.17029 | 0.17183 | – | – |
| EarlyStop LightGBM v2 | 0.17024 | 0.16187 | 0.16187 | 0.18336 |
| Two-stage v1 | 0.17010 | 0.17116 | 0.16173 | 0.18283 |
| LightGBM v1 | 0.17005 | 0.17165 | – | – |
| Multi-Modal EdgeNext | 0.15892 | 0.17410 | 0.16082 | 0.17481 |

**Table 4.** ROC AUC of One-Stage and Two-Stage Models

| Model | OOF | Mean |
|---|---|---|
| Two-stage v2 | **0.97234** | **0.97395** |
| Multi-Modal ConvNext | 0.97082 | 0.97244 |
| XGBoost v2 | 0.96826 | 0.96916 |
| XGBoost v1 | 0.96741 | 0.96817 |
| EarlyStop LightGBM v4 | 0.96719 | 0.96765 |
| LightGBM v3 | 0.96599 | 0.96682 |
| LightGBM v4 | 0.96586 | 0.96666 |
| Two-stage v1 | 0.96518 | 0.96600 |
| EarlyStop LightGBM v2 | 0.96502 | 0.96605 |
| EarlyStop LightGBM v1 | 0.96491 | 0.96629 |
| LightGBM v1 | 0.96485 | 0.96632 |
| Multi-Modal EdgeNext | 0.95110 | 0.96701 |

**Table 5.** Top 15 Retrieval Sensitivity of One-Stage and Two-Stage Models

| Model | OOF | Mean |
|---|---|---|
| Multi-Modal ConvNext | **0.76081** | **0.75995** |
| Two-stage v2 | 0.74809 | 0.75375 |
| XGBoost v2 | 0.73791 | 0.73919 |
| LightGBM v1 | 0.73537 | 0.73769 |
| EarlyStop LightGBM v2 | 0.72774 | 0.73095 |
| LightGBM v4 | 0.72774 | 0.72939 |
| XGBoost v1 | 0.72519 | 0.72621 |
| Two-stage v1 | 0.72265 | 0.72304 |
| EarlyStop LightGBM v4 | 0.72265 | 0.72658 |
| EarlyStop LightGBM v1 | 0.72010 | 0.72296 |
| LightGBM v3 | 0.71501 | 0.71614 |
| Multi-Modal EdgeNext | 0.71247 | 0.71542 |

### Metrics of Three-stage System

Three-stage systems outperform both standalone and two-stage models on the private dataset (Table 6). For this comparison, validation results may not be fully reliable; however, we can evaluate performance based on results from the Private and Public datasets. Finally, compared to the top solutions from the competition (Table 7), our proposed solutions underperform approximately by 2 %. Taking into account a small number of malignant cases, we may consider such a difference to be a marginal one.

Compared to prior works on melanoma detection (Table 8), our solutions outperform all previous approaches. While this comparison is not entirely equitable due to differences in training and validation datasets, most prior studies rely on higher-quality dermoscopic images. Given the lower quality of photo images in our dataset, poorer results might have been expected. However, the findings strongly indicate that our proposed solution performs comparably, if not better, on lower-quality photo images, demonstrating its robustness and applicability in real-world scenarios.

**Table 6.** Metrics of Three-stage Systems

| System | Part ROC AUC | Private | Public | Sensitivity |
|---|---|---|---|---|
| Two-stage v2 | 0.17666 | 0.16941 | **0.18608** | 0.74809 |
| Two-stage v1 | 0.17010 | 0.16173 | 0.18283 | 0.72265 |
| Three-stage v2 | **0.18068** | **0.17042** | 0.18528 | **0.78371** |
| Three-stage v1 | 0.18014 | 0.16982 | 0.18449 | 0.77608 |
| Three-stage v1 mc | 0.17939 | 0.17039 | 0.18527 | 0.78117 |

**Table 7.** Comparison with Best Competition Results

| Solution | Private | Public |
|---|---|---|
| 1st Private Place | **0.17264** | 0.18611 |
| 1st Public Place | 0.17051 | **0.188** |
| Three-stage v1 mc | 0.17039 | 0.18527 |
| Three-stage v2 | 0.17042 | 0.18528 |

**Table 8.** Comparison with Other SOTA Researches

| Experiment | Dataset | ROC AUC |
|---|---|---|
| 2020 Best Solution [16] | 2020 ISIC Competition [48] | 0.9490 |
| Saranya N et al. [14] | PH2 [49] | 0.87 |
| Saranya N et al. [14] | Derm7pt [15] | 0.76 |
| Jojoa Acosta et al. [50] | ISIC 2017 Challenge [21] | 0.91 |
| Ours (Multi-Modal Con-vNext) | ISIC 2024 Kaggle Challenge | 0.97244 |
| Ours (XGBoost v2) | ISIC 2024 Kaggle Challenge | 0.96916 |
| Ours (Two-stage v2) | ISIC 2024 Kaggle Challenge | **0.97395** |

### Image and Multi-Modal Models Ablation Study

As a Baseline model, EfficientNet B1 [51] is used. It uses image resolution 128, severe data augmentations, and a balanced data sampler (described in Section Detailed Neural Net Architecture and Training Setup). As shown in Table 9, the model benefits from adding ISIC Archive data, even considering the modality shift. Regarding increasing image resolution, we observe reasonable score improvements on the Validation set but controversial results on Public and Private sets. Higher resolution noticeably slows down model training and inference, leading us to drop this feature. Discussing different loss functions, such as Focal [52] and ASL [53], we see improvements across all scores, which is expected given the high label imbalance. However, Balanced MixUp [54] does not prove effective for this task. Regarding backbone architecture search across EfficientNet B1, EfficientNet V2 B0, EdgeNeXt Base, and ConvNext V2 Pico, Table 9 shows that the EdgeNeXt family performs the worst, while EfficientNet V2 achieves the best results. The ConvNext family performs in the middle, marginally underperforming EfficientNet V2.

**Table 9.** Ablation Study of Image-Only Model

| **Setup** | **Mean Partial ROC AUC** | **Public** | **Private** |
|---|---|---|---|
| Baseline | 0.15252 | 0.14769 | 0.13588 |
| Add ISIC Archive data | 0.15478 | 0.15500 | 0.14390 |
| Add ISIC Archive data + Resolution 256 | 0.15761 | 0.15208 | 0.14633 |
| Add ISIC Archive data + Focal loss | 0.15889 | 0.15361 | **0.14645** |
| EfficientNet V2 B2 + Add ISIC Archive data | **0.162993** | 0.15020 | 0.13812 |
| ConvNext V2 Pico + Add ISIC Archive and Generated data | 0.15723 | 0.15490 | 0.14388 |
| ConvNext V2 Pico + Add ISIC Archive and Generated data + Balanced MixUp | 0.154894 | 0.14902 | 0.14315 |
| ConvNext V2 Pico + Add ISIC Archive and Generated data + ASL loss | 0.159298 | 0.15581 | 0.14474 |
| EdgeNeXt Base + Add ISIC Archive and Generated data | 0.160584 | 0.14729 | 0.13514 |
| EfficientNet V2 B2 + 2 stage with tune on main data | 0.159025 | **0.15917** | 0.14224 |

All multi-modal setups are first trained in image-only mode on Main, ISIC Archive, and Generated data and then tuned on Main data with tabular features (Sections Multi-Modal Neural Net: Image + Tabular data and Detailed Neural Net Architecture and Training Setup). As shown in Table 10, ConvNext outperforms EfficientNet v2. ASL loss shows the best performance on Validation and Private datasets. Finally, all multi-modal models outperform images only by a significant margin. This is evident because their input information is enhanced with tabular features, which are much less noisy than images.

**Table 10.** Ablation Study of Multi-Modal Image/Tabular Models

| Setup | Mean Partial ROC AUC | Public | Private |
|---|---|---|---|
| ConvNext V2 Pico | 0.17698 | **0.17740** | 0.16409 |
| EfficientNet V2 B0 | 0.16836 | 0.16698 | 0.15547 |
| EfficientNet V2 B2 | 0.16600 | 0.16322 | 0.15908 |
| EdgeNeXt Base | 0.17410 | 0.17481 | 0.16082 |
| ConvNext V2 Pico + Balanced MixUp | 0.17582 | 0.16851 | 0.15337 |
| ConvNext V2 Pico + ASL loss | **0.17740** | 0.17403 | **0.16458** |

### Tabular Models Ablation Study

As a baseline model, LightGBM is utilized, incorporating 85 initial features. As shown in Table 11, the model is further enhanced by adding 42 features based on lesions' spatial, color, and physical properties. Performance significantly improves by including 193 features that aggregate and compare lesion characteristics within the same patient or body region.

**Table 11.** Ablation Study of Tabular Model

| Setup | Mean Partial ROC AUC. | Public | Private |
|---|---|---|---|
| LightGBM with basic features | 0.1586 | 0.1693 | 0.1494 |
| LightGBM with additional features | 0.1604 | 0.1707 | 0.1518 |
| LightGBM with aggregated features | **0.1728** | **0.1837** | **0.1643** |

## DISCUSSION

Our proposed method is based on the fusion of image and metadata information in different ways, resulting in a three-stage system of different models.

Three-stage systems effectively leverage the strengths of multi-modal data integration, allowing them to correct errors from earlier stages. All three-stage systems outperform their two-stage counterparts, even when the first version of the two-stage system (Two-stage v1) is included as part of the three-stage system. This improvement likely arises from the ability of other models in the system to compensate for and correct errors introduced by Two-stage v1.

An important observation is that the first version of the first-level model is not trained on the Main data. As a result, its predictions for the Main data are affected by domain shift, potentially introducing errors and even confusing the second-stage model. We hypothesize that increasing the number of positive samples available to the second-stage model could mitigate this issue, enabling it to correct better and utilize the first-level model's predictions.

Table 12 provides coefficients for individual models and systems used to construct the three-stage system. The automated coefficient search using Optuna assigns higher coefficients to models demonstrating better performance on the validation dataset. However, there are exceptions; for instance, the Multi-Modal Vision Transformer (Multi-Modal EdgeNext) receives a notable coefficient despite not being the top performer. This is likely due to its contribution to system diversity, as it relies on an attention mechanism for decision-making.

**Table 12.** Coefficients for 3-Stage System

| Model | 3-Stage v1 | 3-Stage v1 mc | 3-Stage v2 |
|---|---|---|---|
| Multi-Modal ConvNext | **0.30939** | 0.2 | 0.23926 |
| Two-stage v2 | – | – | **0.25295** |
| XGBoost v2 | 0.18841 | 0.15 | 0.12442 |
| LightGBM v1 | 0.01507 | – | 0.01548 |
| EarlyStop LightGBM v2 | 0.01673 | – | 0.01518 |
| LightGBM v4 | 0.02799 | – | 0.00984 |
| XGBoost v1 | 0.03195 | 0.05 | 0.03067 |
| Two-stage v1 | 0.03730 | **0.25** | – |
| EarlyStop LightGBM v4 | 0.10580 | – | 0.06188 |
| LightGBM v3 | 0.09203 | 0.15 | 0.11032 |
| Multi-Modal EdgeNext | 0.17531 | 0.15 | 0.14000 |

Another noteworthy observation is the comparable performance of three-stage v1 and three-stage v1 mc. This occurs because the two-stage v1 model is re-weighted in the second three-stage system. This re-weighting validates the hypothesis that other models within the system effectively correct the errors introduced by two-stage v1, resulting in comparable or even slightly better performance for three-stage v1 mc. This highlights the value of leveraging diverse models in multi-stage systems to enhance overall robustness and accuracy.

When discussing tabular models and the ablation study, the proposed two-stage feature engineering process shows a clear performance improvement. Another notable observation is that XGBoost outperforms LightGBM, supporting the hypothesis that while LightGBM is more suitable for rapid prototyping, other boosting approaches, such as XGBoost, should be utilized to achieve the best performance.

Regarding the comparison of Tabular and Multi-Modal Vision approaches, interesting trends emerge:

- Multi-Modal ConvNext outperforms Tabular approaches on the validation dataset.
- Tabular approaches outperform Multi-Modal ConvNext on the Public dataset.
- Both approaches exhibit comparable performance on the Private dataset, with a slight preference for Tabular approaches.

These differences may stem from two key factors:

- The small number of malignant tiles in the evaluation datasets introduces noise in the evaluation procedure.
- Variations in data quality between Tabular and Image data originate from different clinics and institutions. For instance, validation folds may contain higher-quality images than the Public dataset. Additionally, the automated feature extraction algorithm likely performs differently depending on the quality of the input images.

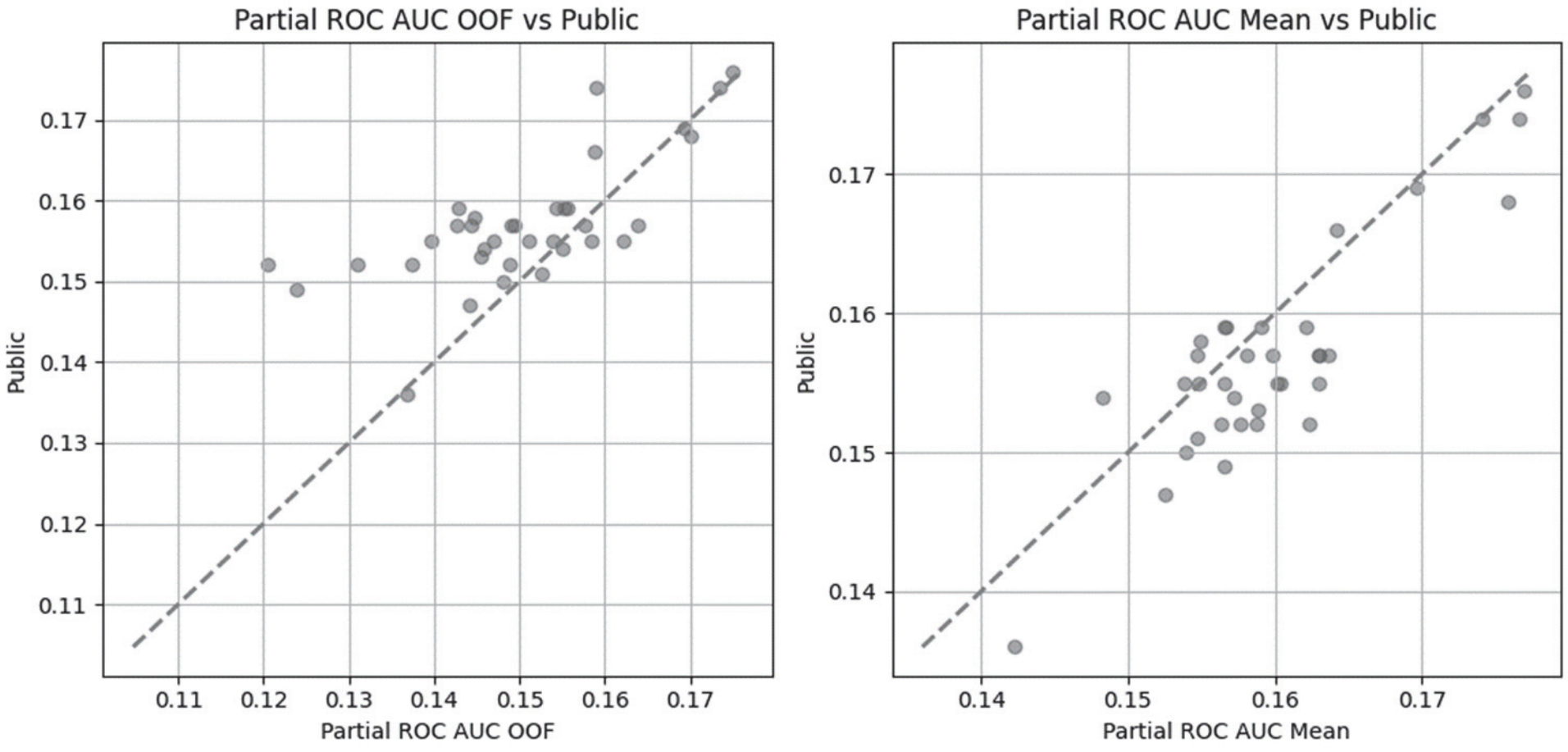

*Fig. 12.* Plot of scores on the validation set and Public set across 39 different Image-only and Multi-Modal experiments

The multi-modal design demonstrates robustness, but the scarcity of datasets combining image and metadata remains a significant limitation. Additionally, variations in data quality across datasets introduce evaluation noise, as reflected in discrepancies between validation and public metrics. In Figure 12, we explore the correlation between scores obtained from two evaluation datasets, using OOF and Mean score for the validation dataset. We can conclude that for both computation approaches (Mean and OOF), the score difference is reasonable for Public and validation sets. However, the Mean score correlates better.

For future work, we identify the following key directions:

– Conduct additional benchmarks on new skin cancer datasets that include metadata to evaluate system generalizability.

– Develop hybrid models capable of aggregating information across multiple nearby lesions to improve classification accuracy.

– Address the domain shift problem between dermoscopic and photo images to bridge the gap between clinical and real-world applications.

## CONCLUSIONS

This paper proposes a three-stage system that leverages multi-modal data, including images and metadata, to classify skin cancer. Unlike previous works, our approach incorporates metadata directly related to the characteristics of individual lesions. We achieve enhanced system performance by employing multiple datasets (both with and without metadata), implementing a multi-step feature engineering pipeline, and using advanced techniques for optimizing performance on highly imbalanced datasets. The experiments are conducted on a novel skin cancer classification dataset composed of photo images, demonstrating the potential applicability of our approach in real-world scenarios, benefiting many patients.

## ACKNOWLEDGEMENTS

First and foremost, we express our deepest gratitude to the Armed Forces of Ukraine, the Security Service of Ukraine, the Defence Intelligence of Ukraine, and the State Emergency Service of Ukraine for ensuring the safety and security that made it possible to complete this work. We also sincerely thank the Kaggle team, Canfield Scientific, The Shore Family Foundation, and all contributing institutions for providing the essential data and materials that enabled us to build models, test hypotheses, and complete this research. The authors acknowledge the use of OpenAI's ChatGPT for text refinement during the preparation of this manuscript. This tool enhanced the text's clarity and flow while ensuring the technical content's accuracy remained intact.

## INFORMATION ON THE ARTICLE

**Volodymyr S. Sydorskyi,** ORCID: 0000-0001-9697-7403, National Technical University of Ukraine "Igor Sikorsky Kyiv Polytechnic Institute", Ukraine, e-mail: volodymyr.sydorskyi@gmail.com

**Igor E. Krashenyi,** ORCID: 0000-0003-0424-147X, Ukrainian Catholic University, Ukraine, e-mail: igor.krashenyi@ucu.edu.ua

**Oleksii P. Yakubenko,** ORCID: 0009-0009-5752-4546, "Pleso Therapy", Ukraine, e-mail: yakubenko.oleksii@gmail.com

**МУЛЬТИМОДАЛЬНА СИСТЕМА ДЛЯ ВИЯВЛЕННЯ МЕЛАНОМИ ШКІРИ /** В.С. Сидорський, І.Е. Крашений, О.П. Якубенко

**Анотація.** Виявлення меланоми є надзвичайно важливим для ранньої діагностики та ефективного лікування. Хоча глибокі нейронні мережі на основі дермоскопічних зображень показали багатообіцяльні результати, їх використання потребує спеціалізованого обладнання, що обмежує їх застосування в ширших клінічних умовах. Представлено мультимодальну систему виявлення меланоми, яка використовує звичайні фотозображення, що робить її більш доступною й універсальною. Ця система інтегрує зображення із табличними метаданими, такими як демографічні дані пацієнтів і характеристики утворів, для підвищення точності виявлення. Вона використовує мультимодальну нейронну мережу, яка поєднує оброблення зображень і метаданих, і підтримує двохетапну модель для випадків із метаданими або без них. Трирівнева система додатково вдосконалює прогнози за допомогою алгоритмів градієнтного бустингу, покращуючи результати. Для вирішення проблем, пов'язаних із дуже незбалансованим набором даних, реалізовано спеціальні техніки, які забезпечують надійне навчання. У дослідженні впливу компонентів оцінено сучасні архітектури комп'ютерного зору, алгоритми бустингу і функції втрат із досягненням пікового значення часткової AUC ROC 0.18068 (максимум 0.2) та чутливості в топ-15 пошуку 0.78371. Результати демонструють, що інтеграція фотозображень із метаданими у багаторівневу систему забезпечує суттєве покращення продуктивності. Ця система просуває виявлення меланоми, пропонуючи масштабоване рішення, яке не залежить від спеціалізованого обладнання і підходить для різноманітних умов надання медичної допомоги, об'єднуючи спеціалізовану та загальноклінічну практики.